\documentclass[british]{llncs}
\usepackage{array}
\usepackage{amssymb}
\usepackage{amsmath}
\usepackage{graphicx}
\usepackage[authoryear]{natbib}
\usepackage{babel}

\begin{document}

%\title{Algorithmic Intelligence Quotient:\\A Practical Measure of Machine Intelligence}
%\title{A Practical Measure of Machine Intelligence}
%\title{On Approximating the\\Universal Intelligence Measure}
%\title{AIQ: An Approximation of the\\Universal Intelligence Measure}
\title{An Approximation of the\\Universal Intelligence Measure}

\iffalse
\author{Shane Legg  \and Joel Veness}
\institute{DeepMind Technologies Ltd \\ \texttt{shane@deepmind.com} \\  University of Alberta \texttt{veness@cs.ualberta.ca}}
\fi

\author{
Shane Legg\\
DeepMind Technologies Ltd\\
\texttt{shane@deepmind.com}\\
\vspace{1em}
Joel Veness\\
University of Alberta\\
\texttt{veness@cs.ualberta.ca}
}
\institute{}
%\institute{DeepMind Technologies Ltd \\ \texttt{shane@deepmind.com} \\  University of Alberta \texttt{veness@cs.ualberta.ca}}

\maketitle

\begin{abstract}
The \emph{Universal Intelligence Measure} is a recently proposed
formal definition of intelligence.  It is mathematically specified,
extremely general, and captures the essence of many informal
definitions of intelligence.  It is based on Hutter's Universal
Artificial Intelligence theory, an extension of Ray Solomonoff's
pioneering work on universal induction.  Since the Universal
Intelligence Measure is only asymptotically computable, building a
practical intelligence test from it is not straightforward.  This
paper studies the practical issues involved in developing a real-world
UIM-based performance metric.  Based on our investigation, we develop
a prototype implementation which we use to evaluate a number of
different artificial agents.
\end{abstract}

\section{Introduction}

A fundamental problem in \emph{strong} artificial intelligence is the
lack of a clear and precise definition of intelligence itself.  This
makes it difficult to study the theoretical or empirical aspects of
broadly intelligent machines.  Of course there is the well-known 
Turing Test \citep{Turing:50}, however this paradoxically seems to be
more about dodging the difficult problem of explicitly defining
intelligence than addressing the real issue.  We believe that
until we have a more precise definition of intelligence, the quest for
generally intelligent machines will lack reliable techniques for
measuring progress.

One recent attempt at an explicit definition of
intelligence is the \emph{Universal Intelligence Measure}
\citep{Legg:07ior}.  This is a mathematical, non-anthropocentric definition of intelligence
that draws on a range of proposed informal definitions of
intelligence, algorithmic information theory \citep{Li:08},
Solomonoff's model of universal inductive inference
\citep{Solomonoff:64,Solomonoff:78}, and Hutter's AIXI theory of
universal artificial intelligence
\citep{Hutter:01aixi,Hutter:04uaibook}.  This paper conducts a preliminary investigation into the
potential for this particular measure of intelligence to serve as a practical metric for evaluating
real-world agent implementations.

\section{Background}

We now briefly describe the recently introduced notion of a
Universal Intelligence Test, the Universal Intelligence Measure
and the practical issues that arise when attempting to evaluate the
performance of broadly intelligent agents.

\subsection{Universal Intelligence Tests}

\cite{Hern10} introduce the notion of a \emph{Universal Intelligence
  Test}, a test designed to be able to quantitatively assess the
performance of artificial, robotic, terrestrial or even
extra-terrestrial life, without introducing an anthropocentric bias.
Related discussion on the motivation behind such tests is given by \cite{DoweHajek98,HernandezOrallo00a,Schaul:10}. 
With respect to our goal of wanting to build more
powerful artificial agents, we strongly support the introduction of such 
general purpose tests.  Having a suite of such tests, with each emphasizing
different, measurable aspects of intelligence, would clearly help the community
build more powerful and robust general agents.  This paper introduces our own such
test, which works by approximating the Universal Intelligence Measure.

\subsection{Universal Intelligence Measure\label{sec:Universal-intelligence-measure}}

After surveying some 70 informal definitions of intelligence proposed by various
psychologists and artificial intelligence researchers, Legg and Hutter \citeyearpar{Legg:07ior} argue
that the informal definition:
\begin{quotation}
\emph{ ``intelligence measures an agent's ability to achieve goals in
  a wide range of environments'', }
\end{quotation}
broadly captures many important properties associated with intelligence.
To formalise this intuition, they used
\emph{reinforcement learning} \citep{Sutton:98}, a general framework
for goal achieving agents in unknown environments.  In this setting,
cycles of interaction occur between the agent and the environment.  At
each cycle, the agent sends an action to the environment, that then 
responds with an observation and (scalar) reward.
The agent's goal is to choose its actions, based on its previous observations and rewards, so as to
maximise the rewards it receives over time. With a little imagination,
it is not hard to see that practically \emph{any} problem can be expressed in
this framework, from playing a game of chess to writing an
award-winning novel.

In their setup, both the agent and environment are expressed as conditional
probability measures over interaction sequences. To formalise a `wide
range of environments', the set of all Turing computable environments
is used, with the technical constraint that the sum of returned
rewards is finitely bounded. Finally, the agent's performance over
different environments is then aggregated into a single
result. To encourage agents to apply Occam's Razor, as advocated by \cite{Legg:07ior},
each environment is weighted according to its complexity, with simpler environments being weighted more heavily.
This is elegantly achieved by using the algorithmic prior distribution \citep{Li:08}. 
The universal intelligence of an agent $\pi$ can then
be defined as,
\begin{equation}
\Upsilon(\pi)\,:=\,\sum_{\mu\in E}2^{-K(\mu)}V_{\mu}^{\pi}\label{eq:ui}
\end{equation}
where $\mu$ is an environment from the set $E$ of all computable
reward bounded environments, $K(\cdot)$ is the Kolmogorov complexity, and $V_{\mu}^{\pi}:=\mathbb{E}(\sum_{i=1}^{\infty}R_{i})$
is the expected sum of future rewards when agent $\pi$ interacts
with environment $\mu$.

This theoretical measure of intelligence has a range of desirable
properties. For example, the most intelligent agent under this measure is Hutter's
AIXI, a universal agent that converges to optimal
performance in any environment where this is possible for a general
agent \citep{Hutter:04uaibook}. At the other end of the scale, it
can be shown that the Universal Intelligence Measure sensibly orders
the performance of simple adaptive agents. Thus, the measure spans
an extremely wide range of capabilities, from the simplest reactive
agents up to universally optimal agents. Unlike the pass or fail Turing
test, universal intelligence is a continuous measure of performance
and so it is more informative of incremental progress. Furthermore,
the measure is non-anthropocentric as it is based on the fundamentals
of mathematics and computation rather than human imitation. 

The major downside is that the Universal Intelligence Measure is only a theoretical definition, 
%The infinite sum, the selection of reference machine, and the incomputability of the complexity function mean that it 
and is not suitable for evaluating real-world agents directly.

\section{Algorithmic Intelligence Quotient}

The aim of the Universal Intelligence Measure was to define
intelligence in the most general, precise and succinct way possible.
While these goals were achieved, this came at the price of asymptotic
computability.  In this section we will show how a practical measure
of machine intelligence can be defined via approximating this notion.
While we will endeavor to retain the spirit of the Universal
Intelligence Measure, the emphasis of this section will be on practicality rather than
theoretical purity.  We will call our metric the \emph{Algorithmic
  Intelligence Quotient} or AIQ\footnote{IQ was originally a quotient, but is now normalised to a Gaussian.  AIQ is also not a quotient, however we use the name since ``IQ" is well understood to be a measure of intelligence.} for short.

\subsection{Environment sampling}

\iffalse
While the Universal Intelligence Measure sums over all computable
environments, in a practical test we will need to approximate this
by a finite sample. 
\fi

One way to define an Occam's Razor prior is to
use the Universal Distribution \citep{Solomonoff:64}. 
The universal prior probability, with respect to a reference machine $\mathcal{U}$, of a sequence beginning with a finite
string of bits $x$ is defined as

\[
M_{\mathcal{U}}(x)\,:=\!\sum_{p:\mathcal{U}(p)=x*}\!\!2^{-\ell(p)},
\]
where $\mathcal{U}(p)=x*$ means that the universal Turing machine
$\mathcal{U}$ computes an output sequence that begins with $x$ when it
runs program $p,$ and $\ell(p)$ is the length of $p$ in bits.  As
the Kolmogorov complexity of $x*$ is the length of the shortest
program for $x*$, by definition, it follows that the largest term in
$M$ is given by $2^{-K(x*)}$. Thus, the set of all sequences that
begin with a low complexity string will have a high prior probability
under $M$, in accordance with Occam's Razor. The difference now
is that the lengths of all programs that generate strings beginning with
$x$ are used to define the prior, not just the shortest program.

The advantage of switching to this related distribution is that it is
much easier to sample from. As the probability of sampling a program
$p$ by uniformly sampling consecutive bits is $2^{-\ell(p)}$, to
sample a sequence from $M$ we just randomly sample a program $p$ and
run it on $\mathcal{U}$. 
% this is not quite true as not all strings are valid programs
This method of sampling has been used to create the test data
sequences that make up the Generic Compression Benchmark
\citep{Mahoney:08}.  Here we will use this technique to sample
environments for the Universal Intelligence Measure. More precisely,
having defined a prefix-free universal Turing machine ${\normalcolor
  \mathcal{U}}$, we generate a finite sample of $N$ programs
$S:=p_{1},p_{2},\ldots,p_{N}$ by uniformly generating bits until we
reach the end of each program.  This is not a set as the same program
can be sampled many times. The estimate of agent $\pi$'s universal
intelligence is then,
\[
\hat{\Upsilon}(\pi)\,:=\,\frac{1}{N}\sum_{i=1}^{N}\hat{V}_{p_{i}}^{\pi},
\]
where we have replaced the expectation $V_{\mu}^{\pi}$ with
$\hat{V}_{p_{i}}^{\pi}$ which is defined to be the empirical total
reward returned from a single trial of environment
$\mathcal{U}(p_{i})$ interacting with agent $\pi$. Since we are 
sampling the space of \emph{programs} that define environments, rather than
the space of environments directly, multiple programs
can define the same environment. Notice that the weighting by
$2^{-\ell(p_{i})}$ is no longer needed as the probability of a program
being sampled decreases by $\frac{1}{2}$ for every additional bit. 
The natural idea of performing a Monte Carlo sample over environments is also used by \cite{Hern10} and \cite{Schaul:10} in their related work.

\subsection{Environment simulation\label{sec:Environment-simulation}}

We need to be able to run each sampled program on our reference machine
$\mathcal{U}$.  A technical problem we face is that some programs will
not halt, and due to the infamous halting problem, we know there is no process 
that can always determine when this is the case.  The
extent of this problem can be reduced by choosing a reference machine
where non-halting programs are relatively unlikely, or one which aids 
the detection of many non-halting programs. Even so, we would still
have non-halting problems to deal with.

From a practical perspective there is not much difference between a
program that does not halt and one that simply runs for too long: in
both cases the program needs to be discarded.  To determine if this is
the case, we first run the program on the reference machine.  If the program exceeds our 
computation limit in any cycle, the program is discarded.
In the future, more powerful hardware will allow us to increase this limit to obtain more accurate AIQ estimates.

\iffalse
Another issue is that the number of computation steps used by the
environment may vary between interaction cycles. One approach, taken
by \citet{HernandezOralloDowe10}, is to define the time to be the maximal
number of computation steps that the environment takes in any cycle.
The problem with this is that in general we never know what the maximal
value is. For example, we may have an environment that computes its
output in just 10 steps for the first million cycles, and then takes
$10^{100}$ steps for all the rest.

Perhaps a more natural approach would be to assume that the
environment runs for a constant number of steps during each cycle. This
would allow us to dodge the halting problem entirely, as we would just need to
read the program's output periodically.  For the reference
machines that we have tested it causes problems, mainly because a
simple program can produce a complex sequence of outputs due to it
being interrupted at different points in its computation during each
cycle. As such we have used a very simple approach:~if a program
exceeds our computation limit in any cycle during a trial, that
program and trial are discarded.
\fi

\subsection{Temporal preference}

In the Universal Intelligence Measure, the total reward that an
environment can return is upper bounded by one.  Because all
computable environments that respect this constraint are considered,
in effect the Universal Intelligence Measure considers all computable
distributions of rewards.  Theoretically this is elegant, but
practically we have no way of knowing if a program will respect the
bound.

A more practical alternative is \emph{geometric discounting}
\citep{Sutton:98} where we allow the environment to generate any
reward in any cycle so long as the reward belongs to a fixed bounded
interval. Rewards are then scaled by a factor that decreases
geometrically with each interaction cycle. Under such a scheme the
reward sum is bounded and thus we can bound the remaining reward left
in a trial. For example, we can terminate each trial once the possible
remaining reward drops below a certain value.

While this is elegant, it is not very computationally efficient when
we are interested in learning over longer time frames.
This is since the later cycles, where the agent has most likely learnt 
the most, are the most heavily discounted.  Thus,
we will focus here on undiscounted, bounded rewards over fixed length trials.

\subsection{Reference machine selection\label{sec:Reference-machine-selection}}

When looking at converting the Universal Intelligence Measure into a
concrete test of intelligence, a major issue is the choice of a suitable
reference machine.  Unfortunately, there is no such
thing as a canonical universal Turing machine, and the choice that we
make can have a significant impact on the test results.  Very powerful agents such as AIXI will achieve high universal intelligence no
matter what reference machine we choose, assuming we allow agents to train from samples prior to taking the test, as suggested in \cite{Legg:07ior}. For more limited agents however, the choice of reference machine
is important.  Indeed, in the worst case it can cause
serious problems~\citep{Hibbard:09}.  When used with typical modern
reinforcement learning algorithms and a fairly natural reference
machine, we expect the performance of the test to lie between these
two extremes.  That is, we expect that the reference machine will be
important, but perhaps not so important that we will be unable to
construct a useful test of machine intelligence.  Providing some
empirical insight into this is one of the main aims of this paper.

Before choosing a reference machine, it is worth considering, in broad terms,
the effect that different reference machines will have on the
intelligence measure.  For example, if the reference machine is like the Lisp
programming language, environments that can be compactly described using
lists will be more probable. This would more heavily weight
these environments in the measure, and thus if we were trying to
increase the universal intelligence of an agent with respect to this particular 
reference machine, we would progress most rapidly if we focused our effort on our
agent's ability to deal with this class of environments. On the other hand, with a
more Prolog like reference machine, environments with a logical rule structure would be more important.
More generally, with a simple reference machine, learning to deal with
small mathematical, abstract and logical problems would be emphasised as
these environments would be the ones computed by small programs. These
tests would be more like the sequence prediction and logical puzzle
problems that appear in some IQ tests. 

\iffalse
\begin{figure}[t!]
\center \includegraphics[scale=0.5]{uimutm.eps}\caption{Measures of
  universal intelligence based on different reference machines.
  Results will differ substantially for weak agents, while more
  powerful agents will converge towards the performance of the AIXI
  agent. \label{Flo:test-range}}
\end{figure}
\fi

What about very complex reference machines?  This would
permit all kinds of strange machines, potentially causing the most
likely environments to have bizarre structures.  As we would like
our agents to be effective in dealing with problems in the real world,
if we do use a complex reference machine, it seems the best choice would be to use
a machine that closely resembles the structure of the real world.  Thus, the
Universal Intelligence Measure would become a simulated version of
reality, where the probability of encountering any given challenge
would reflect its real world likelihood. Between these extremes, a
moderately complex reference machine might include three dimensional
space and elementary physics.  \iffalse We illustrate this spectrum in
Figure~\ref{Flo:test-range}, showing the development paths that one
might follow when optimising an agent with respect to different
Universal Intelligence Measures. \fi While complex reference machines
allow the intelligence measure to be better calibrated to the real
world, they are far more difficult to develop.  Thus, at least for our first set of
tests, we focus on using a very simple reference machine.

\subsection{BF reference machine}\label{sec:bf_ref}

One important property of a reference machine is that it should be
easy to sample from.  The easiest languages are ones where
all programs are syntactically valid and there is a unique end of
program symbol.  One language with this feature is Urban M\"uller's BF
language.  It has just 8 symbols, listed in
Table~\ref{tab:Standard-BF-program} along with their C equivalents,
where we have used C \texttt{stdin} and \texttt{stdout} at the input
and output tapes, and \texttt{p} is a pointer to the work tape.

\begin{table}[t!]
\center
\begin{tabular}{|c|c|c|}
\hline 
BF &  & C\tabularnewline
\hline 
\hline 
\texttt{>} & move pointer right & \texttt{p++;}\tabularnewline
\hline 
\texttt{<} & move pointer left & \texttt{p-{}-;}\tabularnewline
\hline 
\texttt{+} & increment cell & \texttt{{*}p++;}\tabularnewline
\hline 
\texttt{-} & decrement cell & \texttt{{*}p-{}-;}\tabularnewline
\hline 
\texttt{.} & write output & \texttt{putchar({*}p);}\tabularnewline
\hline 
\texttt{,} & read input & \texttt{{*}p = getchar();}\tabularnewline
\hline 
\texttt{{[}} & if cell is non-zero, start loop & \texttt{while({*}p) \{}\tabularnewline
\hline 
\texttt{{]}} & return to start of loop & \texttt{\}}\tabularnewline
\hline 
\end{tabular}
\vspace{1em}
\caption{\label{tab:Standard-BF-program}Standard BF program symbols along with their C equivalents.}
\end{table}

To convert BF for use as a reference machine the agent's action
information is placed on input tape cells, then the program is run,
and the reward and observation information is collected from the
output tape.  
Reward is the first symbol on the output tape and is normalised to the range -100 to +100.  The following symbol is the observation.  All symbols on the input, output and work tapes are integers, with a modulo applied to deal with under/over flow conditions.
As discussed in
Section~\ref{sec:Environment-simulation}, we set a time limit for the
environment's computation in each interaction cycle, here 1000
computation steps. To encourage programs to terminate, we interpret
any attempt to write excess reward and observation cells as a signal
to halt computation for that interaction cycle. As a result about 90\%
of programs do not exceed the computation limit and halt with output
for each cycle.

As we do not wish our environments to always be deterministic, we have
added to BF the instruction \texttt{\%} which writes a random symbol
to the current work tape cell. Furthermore, we also place a history of
previous agent actions on the input tape. This solves the problem of
what to do when a program reads too many input symbols, and it also
makes it easier for the environment to compute functions of the
agent's past actions. Finally, after randomly sampling a program we
remove any pointless code, such as {}``\texttt{+-}'',
{}``\texttt{><}'' and {}``\texttt{{[}{]}}''. This produces faster and
more compact programs, and discards the most common type of pointless
infinite loop. We also discard programs that do not contain any
instructions to either read from the input or write to the output.

Finally, the first bit of the program indicates whether the reward
values are negated or not.  By randomly setting this bit, randomly acting
agents have an AIQ of zero, a natural baseline suggested
by~\cite{Hern10}.

\subsection{Variance Reduction Techniques for AIQ Estimation}

Obtaining an accurate estimate of an agent's AIQ using simple
Monte-Carlo sampling can be time consuming. This is due to the
relatively slow rate at which the standard error decays as the number of samples increases, along
with the fact that for many agents, simulating even a single episode is quite demanding.
To help our implementation provide statistically significant results within reasonable time constraints, 
we applied a number of techniques that significantly reduced the variance of our AIQ estimates.

The first technique was to simply exploit the parallel nature of Monte Carlo
sampling so that the test could be run on multiple cores. On present day hardware,
this can easily lead to a 10x performance improvement over a single core implementation.

The second technique was to use \emph{stratified sampling}.
It works as follows: first, the sample space
$\Omega$ is partitioned into $k$ mutually exclusive sets
$\Omega_{1},\Omega_{2},\dots,\Omega_{k}$ such that
$\bigcup_{i=1}^{k}\Omega_{i}=\Omega$. Each $\Omega_{i}$ is called a
\emph{stratum}. The total probability mass $\Pr[X\in\Omega_{i}]$
associated with each of the $k$ strata needs to be known in
advance.  Given a sample $(X_{1},X_{2},\dots,X_{n})$, the stratified
estimate $\hat{X}_{ss}$ is given by,

\[
\hat{X}_{ss}:= \sum_{i=1}^{k}\Pr[X\in\Omega_{i}]\left(\frac{1}{n_{i}}\sum_{j=1}^{n}X_{j}\mathbb{I}[X_{j}\in\Omega_{i}]\right)
\]
where $n_{k}:=\sum_{i=1}^{n}\mathbb{I}[X_{i}\in\Omega_{k}]$. It can
be interpreted as a convex combination of $k$ simple Monte Carlo
estimates, and is easily shown to be unbiased.
For a fixed sample size, the optimal way to allocate samples is in
proportion to the standard deviation of each stratum, weighted by the
stratum's probability mass.  More precisely, if $f_{X}(x)$ is the
density function of $X$ and $f_{k}(x)\mathbb{\propto
  I}[x\in\Omega_{k}]f_{X}(x)$ is the density function associated with
the random variable $Y_{k}$ associated with stratum $k$, the optimal
allocation ratio is achieved when
$n_{k} \propto \sqrt{\mathrm{Var}[Y_{k}]}\Pr[X\in\Omega_{k}]$.  To do
this we must estimate $\mathrm{Var}[Y_{k}]$ during sampling and adapt
which strata we are drawing samples from accordingly. Intuitively, the
algorithm is identifying those parts of the sample space which have
the most variance and are of the most significance to the final
result, and concentrating the sampling effort in these regions.  There
are various algorithms for adaptive stratified sampling, however we
have chosen the method developed by \cite{Etore:10} as they have
derived the confidence intervals for the estimate of the mean, a feature
we will use when reporting our results. 
In AIQ, we stratified on a combination of simple properties of each
environment program, including the length and the presence of particular patterns of BF symbols.
This particular technique gave roughly a 4x performance increase.

\iffalse
The major practical difficulty is choosing good strata. Ideally, we
would like to divide the program space up into strata such that the
agent's performance within each stratum is relatively homogeneous.
Ideally, we could create strata such at all programs within each stratum
compute the same environment. For example, we might find two programs
that require the agent to choose a specific pattern of actions. Unfortunately,
determining whether two Turing complete programs compute the same
function is a well known undecidable problem \citet{Rice:53}. Nevertheless,
if we simulate an agent that performs random actions in an environment
over a number episodes we can determine the nature of very simple
environments with high accuracy. For example, one of the most common
types of environment is one where the agent's last action is returned
to it as the next reward. Essentially, this is a simple kind of function
maximisation problem over the agent's action space. Another common
pattern is to returns as reward the action that the agent too two
cycles ago. We have identified 10 such categories and used them for
strata. The remaining programs are split across another 10 strata
depending on their program length. Because the process of classify
programs into strata is a little time consuming, we perform this computation
separately and store the results in a file of classified program samples.
\fi

%\subsection{Common random numbers}

Another variance reduction technique we used was \emph{common
  random numbers.}  Rather than estimating the AIQ of two agents $\pi$
and $\pi'$ from independent samples from the environment distribution,
we instead estimate the difference,
\[
\hat{\Delta}(\pi,\pi'):=\hat{\Upsilon}(\pi')-\hat{\Upsilon}(\pi)
\]
using a single set of program samples. 
This technique is particularly important when an agent designer is deciding whether or not to 
accept a new version of the agent.  Intuitively, common random numbers
reduces the chance of one agent performing better due to being
evaluated on an easier sample.  More precisely,
\[
\mathrm{Var}[\hat{\Delta}(\pi,\pi')]=\mathrm{Var}[\hat{\Upsilon}(\pi')]+\mathrm{Var}[\hat{\Upsilon}(\pi)]-2\mathrm{Cov}[\hat{\Upsilon}(\pi'),\hat{\Upsilon}(\pi)].
\]
If independent samples were used for $\hat{\Upsilon}(\pi')$ and
$\hat{\Upsilon}(\pi)$ the covariance would vanish. However, since we
are using a single sample and have assumed that the AIQs of $\pi$ and
$\pi'$ are positively correlated (which makes sense if $\pi'$ is an incremental improvement over $\pi$),
$\mathrm{Cov}[\hat{\Upsilon}(\pi'),\hat{\Upsilon}(\pi)]$ is positive
and thus $\mathrm{Var}[\hat{\Delta}(\pi,\pi')]$ is reduced.

%\subsection{Antithetic Variates}

The final variance reduction technique we used was \emph{antithetic variates}. 
The intuition is quite straightforward:~instead of using one sample, use two samples in such a way 
that the resultant estimators for the first and second sample are negatively correlated. These
can then be combined to balance each other out, thus reducing the total variance. 
More formally, if $\hat{Y}_{1}$and $\hat{Y}_{2}$ are two unbiased estimates of a quantity of interest,
then $\hat{{X}}=\tfrac{1}{2} [ \hat{Y}_{1}+\hat{Y}_{2}]$ is also an unbiased estimator, with
\[
\mathrm{Var}(\hat{{X})}=\tfrac{1}{4} \left[ {\mathrm{Var}(\hat{Y}_{1})}+\mathrm{Var}(\hat{Y}_{2})+2\mathrm{Cov}(\hat{Y}_{1},\hat{Y}_{2}) \right].
\]
Thus if the two estimates are negatively correlated, $\mathrm{Var}(\hat{{X})}$ is reduced.
A common way to achieve this is to sample in pairs, with each element of the pair directly opposing the other in some sense.
In our AIQ implementation, since the first bit of each program specifies whether or not to negate the
rewards, applying antithetic variates was trivial: we simply ran each
program twice, once with the first bit off, once with the first bit on.
This lead to a performance improvement that varied based on the agent being tested.
With the exception of the Random agent (where there was a massive negative correlation), the performance improvements were typically smaller than a factor of 1.5x.

\iffalse
For an agent that makes random actions, flipping the sign of the
reward flips the sign of the expected total reward over a trial. This
produces a strong negative correlation, and thus antithetic variates
helps our estimate of the AIQ of a random agent converge to zero much
more quickly. However, more intelligent agents will adapt to many of
the test environments, and so in many cases if we flip the sign of the
reward the agent will adapt its behavior accordingly. This reduces the
negative correlation and thus weakens the power of this variance
reduction technique for more capable agents.
\fi

\section{Empirical results}

\begin{figure}[t!]
\vspace{-2em}
\includegraphics[scale=0.55]{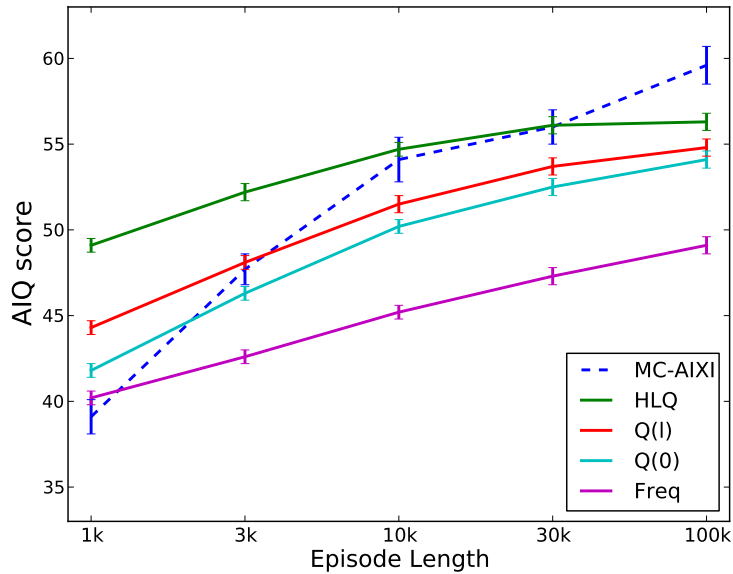}
\caption{Estimated AIQ scores of agents as a function of episode length.\label{Flo:bf5}}
\end{figure}

We implemented AIQ with the variance reduction techniques previously described,
along with the extended BF reference machine.  Our code is open source and available for
download at \texttt{www.vetta.org/aiq}.
%% need to do this!!!  
It should run on any platform containing Python and the Scipy library.
We have also implemented a number of reinforcement
learning agents to test AIQ with.  The simplest agent is called
Random, which makes uniformly random actions. A slightly more complex
agent is Freq, that computes the average reward associated with each
action, ignoring observation information.  It chooses the best action
in each cycle except for a fixed fraction of the time when it tries a
random action.  We have implemented the $Q(\lambda)$ algorithm
\citep{Watkins:89}, which subsumes the simpler $Q(0)$ algorithm as a
special case, and also $HLQ(\lambda)$ which is similar except that it
automatically adapts its learning rate~\citep{Hutter:07hl}.  Finally,
we have created a wrapper for MC-AIXI \citep{Veness:10,veness10b}, a more
advanced reinforcement learning agent that can be viewed as an
approximation to Hutter's AIXI.

\iffalse Before proceeding to the test results, a key point to bear in
mind is that no combination of tests can establish whether the AIQ
measure is in some sense correct. All we can show is that when we test
various agents we get answers that are are not obviously wrong. For
example, if we were to find that a weak agent had a higher
intelligence score than a clearly superior one, this would be
significant evidence that our intelligence measure was
failing. However, if we observe the reverse then all we know is that
in this particular case the intelligence measure is behaving in the
expected way. Perhaps after a great many tests if no significant
inconsistencies are found, this could constitute evidence that the
test appears to be functioning well. Nevertheless, in the absence of a
gold standard for machine intelligence for us to measure our test
against, the ultimate justification that the test indeeds measures
something that can be termed intelligence comes down to the theory and
analysis behind the test.  \fi

\subsection{Comparison of artificial agents}

For our first set of tests we took the BF reference machine and set
the number of symbols on the tape to 5.  We then tested all our agents
without discounting on a range of different episode lengths.  With the
exception of MC-AIXI, which is significantly more computationally expensive, we
performed 10,000 samples in each test.  As expected, the AIQ of the Random agent
was zero.  For the other agents we ran parameter sweeps to find the
best performing settings.  These results appear in
Figure~\ref{Flo:bf5}, with the error bars representing approximate 95\% confidence
intervals.

\iffalse
\begin{figure}[t!]
\hspace{-2.0em} 
\includegraphics[scale=0.34]{BF5.eps}
\hspace{-2.5em} 
\includegraphics[scale=0.34]{BF10.eps}
\caption{AIQ scores across two different BF\label{Flo:bf5}}
\end{figure}
\fi

\begin{figure}[t!]
%\center
\hspace{-2.0em} 
\includegraphics[scale=0.34]{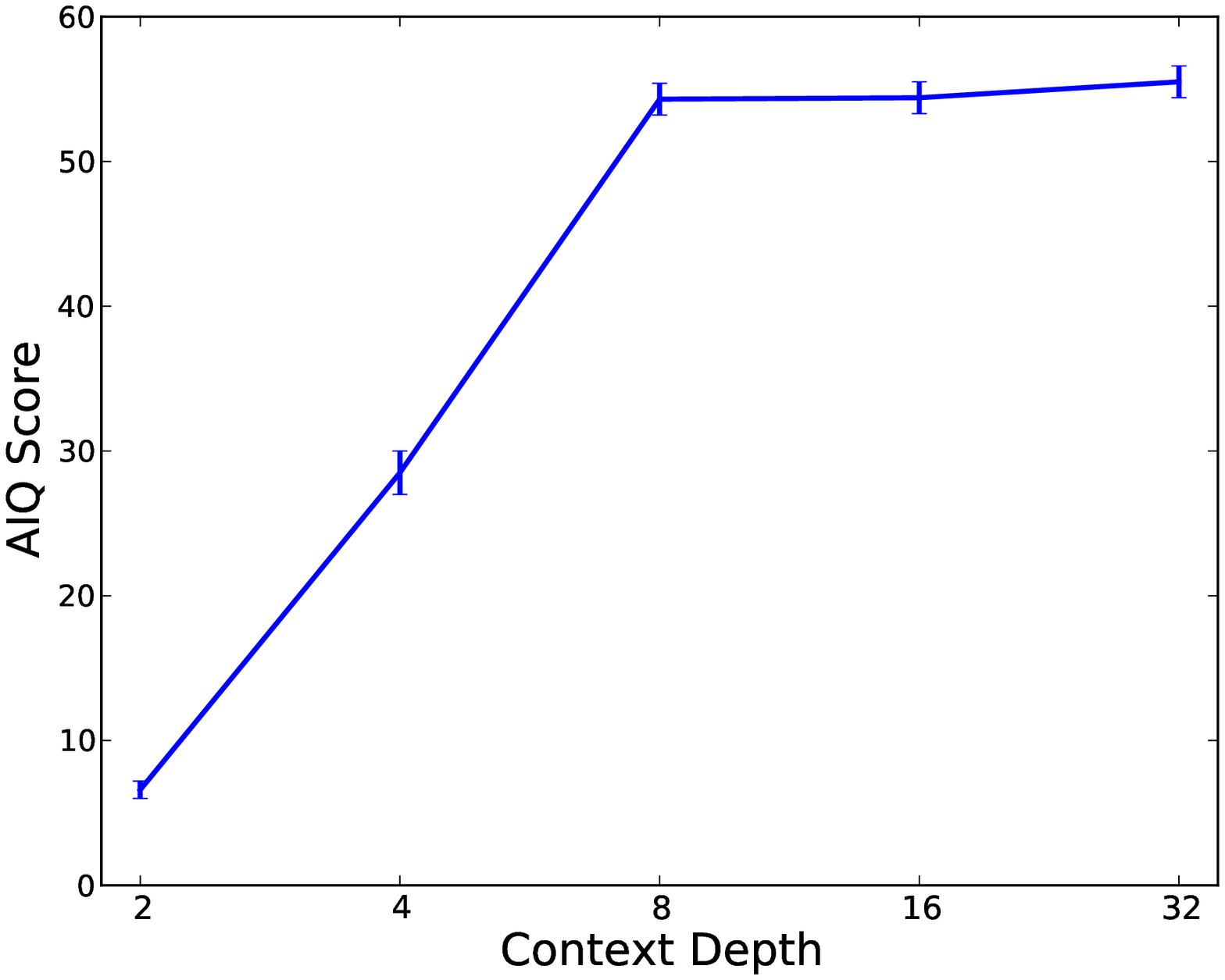}
\hspace{-2.5em}
\includegraphics[scale=0.34]{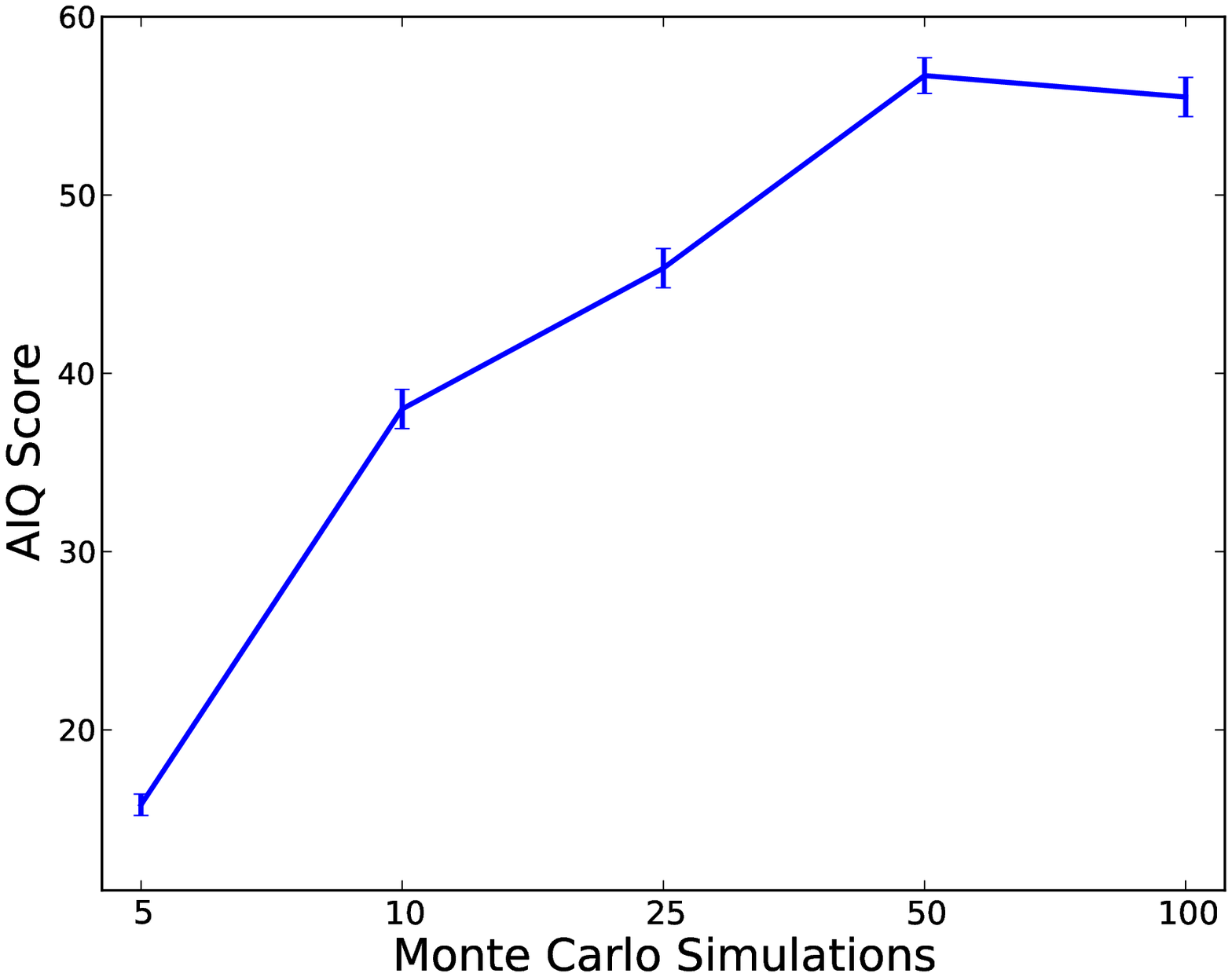}
\caption{Estimated\label{fig:context_depth} AIQ of MC-AIXI as the
  context depth and search effort is varied.}
\end{figure}

For 100k length episodes the agents' AIQ scores appear in the order
that we would expect: Random (not shown), Freq, $Q(0)$, $Q(\lambda)$,
$HLQ(\lambda)$ and MC-AIXI.  As the episode lengths decrease, the
agent's have less learning time in each trial and thus their scores
decline.  Except for MC-AIXI, the relative ranking of the agents remained the same.
It seems MC-AIXI's complex world model is relatively slow to learn but ultimately
the most powerful.  Our initial attempts at modifying
MC-AIXI to be similarly high scoring on shorter runs failed.
Longer tests may be needed in order to determine whether some of the more complicated
agents have reached their maximal AIQ.

Similar tests to the above were performed with 2, 10 and 20 symbol
tapes.  The results were qualitatively the same, but with larger
action and observation spaces the learning times increased for all 
agents.  We also increased the number of cells used to represent the
observations, usually set to 1, which had the same effect.  We
then tried reversing the order of the observation and the reward on the tape, which 
lead to results that were qualitatively the same.  We experimented with
discounting, and the results were consistent with the
undiscounted results using shorter episodes lengths.  We also
increased the computation limit per cycle and did not see any
measurable effect.  Thus our initial findings were that the results seemed 
relatively robust to minor modifications of the reference machine.

\subsection{Measuring Agent Scalability}

The MC-AIXI agent has a parameter that sets the context depth of its
prediction algorithm, in effect controlling the maximal size of the
world model that it can learn.  It also has a parameter that specifies
the number of Monte Carlo simulations it generates, in effect
controlling the amount of effort that it puts into planning for each
interaction cycle.  These two parameters allow us to vary the power of
the MC-AIXI agent along two fundamentally different dimensions.  We
did this with a 5 symbol BF reference machine, as before, and with 50k
length episodes.  The results of these tests appear in
Figure~\ref{fig:context_depth}.  While increasing the agent's search
effort consistently increased its AIQ score, the results for the
context depth appear to have plateaued at a depth of 8, though with
the present error bars it is impossible to tell for sure.  
This warrants future investigation. For example, it may be the case that larger
context depths help only if the episode length is longer than 50,000. 
\iffalse
Alternatively,
the difference could just be really small.
\fi

\subsection{Environment Distribution}

\begin{figure}[t!]
\center
\includegraphics[scale=0.40]{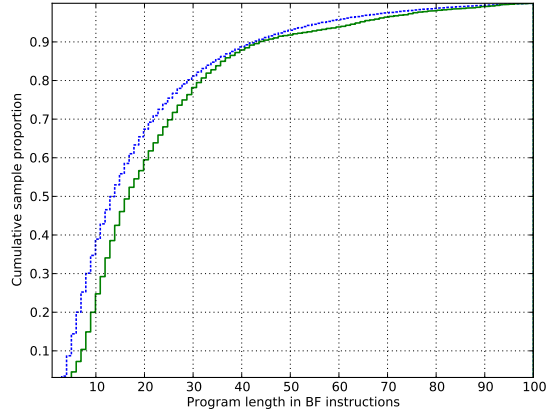}
\caption{\label{fig:env_dist} A comparison of the BF program lengths in the environment distribution compared to the environments chosen by the adaptive sampler. The dashed blue line shows the cumulative proportion of BF environments satisfying a given maximum program length. The solid green line shows the cumulative proportion of BF environments sampled by our variance reduction enhanced adaptive sampling procedure.}
\end{figure}

We next ran some tests to help characterise our environment sampling procedure.

Our first test involved generating $2 \times 10^5$ legal BF environment programs satisfying the criteria listed in Section \ref{sec:bf_ref}.
For example, programs that ran too long or didn't have both a read and a write instruction were discarded. %, which is why there are no programs of length 1 or 0.
The dashed blue line in Figure \ref{fig:env_dist} shows the resultant empirical cumulative distribution of program lengths across the space of BF environments.
Although the number of programs at any given length decays exponentially, this result shows that a significant amount of the total probability mass is still allocated to relatively complex environments with description lengths of 20 symbols or more.

Our next test involved inspecting the distribution of programs sampled by our adaptive sampler when evaluating the HLQ agent.
This is shown by the green line on Figure \ref{fig:env_dist}. 
This shows that the adaptive sampler reduces the proportion of programs of length 10 or less from almost 40\% to 20\%.
On the other hand, from length $20$ to $40$ the green line climbs more quickly then the blue one. 
Thus we see that the adaptive sampler has moved the sampling effort away from programs shorter than 20 symbols, and focused its effort on the $20$ to $40$ symbol range.
%This indicates that the agent's performance was relatively stable for shorter programs, but then become increasing variable.  

We also visually inspected a variety of generated environments.
While it is true that extremely short programs, for example those of less than 5 symbols, do not generate very interesting environments, we found that by the time we got to programs of length 30, many environments (at least to our eyes) seemed quite incomprehensible.
%Thus, we would expect the sweet spot for distinguishing between different moderately powerful agents to be in the 10 to 40 length range that the adaptive sampler has focused its efforts on.
%Might also want to point out that for extremely intelligent agents we would expect to see a more dramatic shift because the agent would perform consistently on all short to medium length programs.

\section{Related Work and Discussion}

\cite{Hernandez-orallo98aformal} developed a related test, called the C-test, that is also based on a
very simple reference machine.  
Like BF it uses a symbolic alphabet with an end wrap around.
Unlike BF, which is a tape based machine, the C-test uses a register machine with just three        
symbol registers.  
This means that the state space for programs is much smaller than in BF.
Another key difference is that the C-test considers generated sequences of symbols, rather than fully interactive       
environments.  
In our view, this makes it not a complete test of intelligence.
For example, the important problem of exploration does not feature in a non-interactive setting.
Extending the C-test reference machine to be interactive would likely be straightforward: simply add instructions to read
and write to input and output tapes, the same way BF does.  
It would be interesting to see how AIQ behaves when using such a reference machine.

A different approach is used by \cite{Insa-etal2011b,Insa-etal2011a}.  Here an interactive reinforcement learning setting is considered, however the space of environments is no longer sampled from a Turing complete reference machine.  Instead a small MDP is used (3, 6 and 9 states) with uniformly random transitions.  Which state is punishing or rewarding follows a fixed random path through this state space.  To measure the complexity of environments, the gzip compression algorithm is applied to a description of the environment.  While this makes the test tractable, in our view it does so in a way that deviates significantly from the Universal Intelligence Measure that we are attempting to approximate with AIQ.  Interestingly, in their setting human performance was not better than the simple tabular Q-learning algorithm.  We suspect that this is because their environments have a simple random pattern structure, something that algorithms are well suited for compared to humans.  
%While our emphasis is on machine intelligence, we did conduct some preliminary tests which suggested that at least the author's AIQs are well ahead of the algorithms tested in this paper.

Another important difference in our work is that we have directly sampled from program space.
This is analogous to the conventional construction of the Solomonoff prior, which samples random
bit sequences and treats them as programs. 
With this approach all programs that compute some environment count towards the
environment's effective complexity, not just the shortest, though the shortest
clearly has the largest impact.  This makes AIQ very efficient in practice since
we can just run sampled programs directly, avoiding the need to have to compute complexity
values through techniques such as brute force program search.  For example, to compute the
complexity of a 15 symbol program, the C-test
required the execution of over 2 trillion programs.  For longer programs,
such as many that we have used in our experiments, this would be completely intractable.  One
disadvantage of our approach, however, is that we never know the complexity
of any given environment; instead we know just the length of one particular program that computes it.

\section{Conclusion}

We have taken the theoretical model of %Universal Intelligence set out by
\citet{Legg:07ior} and converted it into a practical test for
machine intelligence.  To do this we have randomly sampled programs
from a simple universal Turing machine, drawing inspiration at points
from \cite{Hern10}, and the related work in \cite{Hernandez:10}.  In
all of our tests the AIQ scores behaved sensibly, with
agents expected to be more intelligent having higher AIQ.
Naturally, no empirical test can confirm that a test of intelligence
is indeed ``correct'', rather it can only confirm that the theoretical
model behaves as expected when suitably approximated, and that no
insurmountable difficulties arise when attempting this.  We believe
that our present efforts have been successful in this regard, but more work is clearly required.

Perhaps the most worrying potential problem with the Universal
Intelligence Measure is its dependence on the choice of reference
machine, as highlighted by \cite{Hibbard:09}.  While we accept that
problematic reference machines exist, it was our belief that if we
chose a fairly simple and natural reference machine, the resulting
intelligence test would behave sensibly.  While we have only provided
one data point to support this claim here, the fact that it was the first
and only reference machine that we tried gives us hope that it is not
overly special.  Furthermore, we found that the results were
qualitatively the same for a range of minor modifications to the
BF reference machine.  Obviously, further reference
machines will need to be implemented and tested to gain a greater
understanding of these issues.

\subsubsection*{Acknowledgements}
This research was supported by Swiss National Science Foundation grant
number PBTIP2-133701.

%\scriptsize
\makeatletter
\newcommand*\mysize{%
  \@setfontsize\mysize{8}{9.0}%
}
\makeatother
\mysize

\bibliographystyle{apalike}
\bibliography{AIQ}

\begin{thebibliography}{}

\bibitem[Dowe and Hajek, 1998]{DoweHajek98}
Dowe, D.~L. and Hajek, A.~R. (1998).
\newblock {A non-behavioural, computational extension to the {T}uring {T}est}.
\newblock In {\em Intl. Conf. on Computational Intelligence \& multimedia
  applications (ICCIMA'98), Gippsland, Australia}, pages 101--106.

\bibitem[\'Etor\'e and Jourdain, 2010]{Etore:10}
\'Etor\'e, P. and Jourdain, B. (2010).
\newblock Adaptive optimal allocation in stratified sampling methods.
\newblock {\em Methodology and Computing in Applied Probability},
  12(3):335--360.

\bibitem[Hern{\'a}ndez-Orallo, 2000]{HernandezOrallo00a}
Hern{\'a}ndez-Orallo, J. (2000).
\newblock Beyond the {T}uring {T}est.
\newblock {\em J. Logic, Language \& Information}, 9(4):447--466.

\bibitem[Hern{\'a}ndez-Orallo, 2010]{Hernandez:10}
Hern{\'a}ndez-Orallo, J. (2010).
\newblock A (hopefully) unbiased universal environment class for measuring
  intelligence of biological and artificial systems.
\newblock In {\em Proc. of the Third Conference on Artificial General
  Intelligence}, Lugano.

\bibitem[Hern{\'a}ndez-Orallo and Dowe, 2010]{Hern10}
Hern{\'a}ndez-Orallo, J. and Dowe, D.~L. (2010).
\newblock Measuring universal intelligence: Towards an anytime intelligence
  test.
\newblock {\em Artificial Intelligence}, 174(18):1508 -- 1539.

\bibitem[Hernandez-orallo and Minaya-collado, 1998]{Hernandez-orallo98aformal}
Hernandez-orallo, J. and Minaya-collado, N. (1998).
\newblock A formal definition of intelligence based on an intensional variant
  of algorithmic complexity.
\newblock In {\em EIS'98}.

\bibitem[Hibbard, 2009]{Hibbard:09}
Hibbard, B. (2009).
\newblock Bias and no free lunch in formal measures of intelligence.
\newblock {\em Journal of Artificial General Intelligence}, 1(1):54--61.

\bibitem[Hutter, 2001]{Hutter:01aixi}
Hutter, M. (2001).
\newblock Towards a universal theory of artificial intelligence based on
  algorithmic probability and sequential decisions.
\newblock {\em Proc. 12th Eurpean Conference on Machine Learning (ECML-2001)},
  pages 226--238.

\bibitem[Hutter, 2005]{Hutter:04uaibook}
Hutter, M. (2005).
\newblock {\em Universal Artificial Intelligence: Sequential Decisions based on
  Algorithmic Probability}.
\newblock Springer, Berlin.
\newblock 300 pages, http://www.hutter1.net/ai/uaibook.htm.

\bibitem[Hutter and Legg, 2007]{Hutter:07hl}
Hutter, M. and Legg, S. (2007).
\newblock Temporal difference updating without a learning rate.
\newblock In {\em Neural Information Processing Systems (NIPS '07)}.

\bibitem[Insa-Cabrera et~al., 2011a]{Insa-etal2011a}
Insa-Cabrera, J., Dowe, D.~L., Espana-Cubillo, S., Hernandez-Lloreda, M., and
  Hernandez-Orallo, J. (2011a).
\newblock Comparing humans and ai agents.
\newblock In Juergen~Schmidhuber, Kristinn R.~Thorisson, M.~L., editor, {\em
  Artificial General Intelligence, 4th Intl Conf, Mountain View, San
  Francisco}. Lecture Notes in Artificial Intelligence, Springer.

\bibitem[Insa-Cabrera et~al., 2011b]{Insa-etal2011b}
Insa-Cabrera, J., Dowe, D.~L., and Hernandez-Orallo, J. (2011b).
\newblock Evaluating a reinforcement learning algorithm with a general
  intelligence test.
\newblock In Jose A.~Lozano, Jose A.~Gamez, J. A.~M., editor, {\em Current
  Topics in Artificial Intelligence. 14th Conference of the Spanish Association
  for Artificial Intelligence, CAEPIA 2011}. Lecture Notes in Artificial
  Intelligence, Springer.

\bibitem[Legg and Hutter, 2007]{Legg:07ior}
Legg, S. and Hutter, M. (2007).
\newblock Universal intelligence: A definition of machine intelligence.
\newblock {\em Minds and Machines}, 17(4):391--444.

\bibitem[Li and Vit\'anyi, 2008]{Li:08}
Li, M. and Vit\'anyi, P. M.~B. (2008).
\newblock {\em An introduction to {Kolmogorov} complexity and its
  applications}.
\newblock Springer, 3nd edition.

\bibitem[Mahoney, 2008]{Mahoney:08}
Mahoney, M. (2008).
\newblock Generic compression benchmark.
\newblock {\em \mbox{http://www.mattmahoney.net/dc/uiq}}.

\bibitem[{Schaul} et~al., 2011]{Schaul:10}
{Schaul}, T., {Togelius}, J., and {Schmidhuber}, J. (2011).
\newblock {Measuring Intelligence through Games}.
\newblock {\em ArXiv e-prints}.

\bibitem[Solomonoff, 1964]{Solomonoff:64}
Solomonoff, R.~J. (1964).
\newblock A formal theory of inductive inference: Part 1 and 2.
\newblock {\em Inform. Control}, 7:1--22, 224--254.

\bibitem[Solomonoff, 1978]{Solomonoff:78}
Solomonoff, R.~J. (1978).
\newblock Complexity-based induction systems: comparisons and convergence
  theorems.
\newblock {\em IEEE Trans. Information Theory}, IT-24:422--432.

\bibitem[Sutton and Barto, 1998]{Sutton:98}
Sutton, R. and Barto, A. (1998).
\newblock {\em Reinforcement learning: An introduction}.
\newblock Cambridge, MA, MIT Press.

\bibitem[Turing, 1950]{Turing:50}
Turing, A.~M. (1950).
\newblock Computing machinery and intelligence.
\newblock {\em Mind}.

\bibitem[Veness et~al., 2010]{Veness:10}
Veness, J., Ng, K.~S., Hutter, M., and Silver, D. (2010).
\newblock Reinforcement learning via {AIXI} approximation.
\newblock In {\em Proc. Association for the advancesment of artificial
  intelligence}.

\bibitem[Veness et~al., 2011]{veness10b}
Veness, J., Ng, K.~S., Hutter, M., Uther, W., and Silver, D. (2011).
\newblock A {M}onte-{C}arlo {AIXI} {A}pproximation.
\newblock {\em Journal of Artificial Intelligence Research (JAIR)}, 40(1).

\bibitem[Watkins, 1989]{Watkins:89}
Watkins, C. (1989).
\newblock {\em Learning from Delayed Rewards}.
\newblock PhD thesis, King's College, Oxford.

\end{thebibliography}

\end{document}